\documentclass[11pt]{article}

\usepackage{latexsym,verbatim}
\usepackage{amssymb,amsmath,url,color}

\newtheorem{definition}{Definition}[section]
\newtheorem{lem}[definition]{Lemma}
\newtheorem{theorem}[definition]{Theorem}
\newtheorem{corollary}[definition]{Corollary}
\newtheorem{example}[definition]{Example}
\newenvironment{proof}{{\bf Proof:} }{\hfill $\Box$}   

\newcommand{\scheme}{\mathsf{SS}}
\newcommand{\sign}{\mathsf{sig}}
\newcommand{\ver}{\mathsf{ver}}
\newcommand{\cert}{\mathsf{cert}}
\newcommand{\event}{E}
\newcommand{\history}{\mathsf{Hist}}

\begin{document}

\title{Securing emergent behaviour in swarm robotics}

\author{Liqun Chen \\
  Department of Computer Science, \\ University of Surrey, Guildford, Surrey GU2 7XH, United Kingdom. \\ \texttt{liqun.chen@surrey.ac.uk}
    \and 
Siaw-Lynn Ng \\ 
Information Security Group, \\ Royal Holloway, University of London, Egham, Surrey TW20 0EX,  United Kingdom. \\  \texttt{s.ng@rhul.ac.uk}
}

\maketitle

\begin{abstract}
Swarm robotics is the study of how a large number of relatively simple
robots can be designed so that a desired collective behaviour emerges
from the local interactions among robots and between the robots and
their environment.  While many aspects of a swarm may be modelled as
various types of ad hoc networks, and accordingly many aspects of
security of the swarm may be achieved by conventional means, here we
will focus on swarm \emph{emergent behaviour} as something that most
distinguishes swarm robotics from ad hoc networks.  We discuss the
challenges emergent behaviour poses on communications security, and by
classifying a swarm by types of robots, types of communication
channels, and types of adversaries, we examine what classes may be
secured by traditional methods and focus on aspects that are most
relevant to allowing emergent behaviour.  We will examine how this can
be secured by ensuring that communication is secure.  We propose a
simple solution using hash chains, and by modelling swarm
communications using a series of random graphs, we show that this
allows us to identify rogue robots with a high probability.

\end{abstract}

\section{Introduction} \label{sec:intro}

There are many variations on what the term ``swarm robotics'' means
exactly.  From \cite{Sahin,Navarro,HTM,LaingNgTomlinsonMartin} we see
that it is generally agreed that a swarm is a collection of a large
number of autonomous mobile robots. The robots are generally
resource-constrained and have low capability individually.  They are
self-organised, and no sychronicity is assumed. The swarm exhibits
collective emergent behaviour: a desired collective behaviour emerges
from the local interactions among the robots and between the robots
and the environment.

What differs in diverse applications and scenarios are the
communication capabilites, the hierarchical organisation, and the
presence or absence of a central control.  Some swarms consist of
homogeneous robots but some may allow a few groups of homogeneous
robots or allow some robots to take on special roles in communications
or control. There is generally no centralised control, although in
some applications a global channel is needed for a central control to
download information onto the robots, or for robots to report back
findings from the field.  Robots are generally assumed to have local
sensing and communication capabilities but may communicate with their
neighbours in different ways, for example, they may communicate
directly in a peer-to-peer fashion, or they may observe physical
traits of their neighbours, or they may communicate by leaving
messages in the environment.  These variations have impact on the
provision of security to the swarm, and we will discuss them in more
detail in Section \ref{sub:class}.

In \cite{Navarro} basic behaviours of swarm robotics are enumerated,
and in \cite{LaingNgTomlinsonMartin} it is argued that typical
behaviours of swarm robots are either aggregation or dispersion, and
in these behaviours it is the ability to distinguish robots of the
same swarm that is key to the success of the swarm.  This can only be
done if there are some secret common to the swarm that is not
available to adversaries or observers.  In
\cite{LaingNgTomlinsonMartin} the use of public key cryptography and
key predistribution is briefly explored.  We will consider this with
more depth here.


Swarm robotics have applications in many areas, for example, military
(such as mine clearance, access, surveillance), environmental
monitoring, disaster relief and health care (such as medication
provision and monitoring). We refer the reader to \cite{Sahin} for a
more detailed picture.  In many of these applications it is paramount
that the swarm is protected so that the goals can be achieved.  What
``security'' means differ from scenario to scenario, and is discussed
in some detail in \cite{HTM,LaingNgTomlinsonMartin}.  In any case the
desirable properties of a swarm in many applications are common: data
is collected by many robots so the loss of individual robots has
little impact on the success of the task.  This redundancy makes the
swarm reliable.  The distributed nature of the swarm also ensures that
there are no single points of failure.  Any measures taken to increase
security must not adversely affect these properties.

In \cite{HTM, LaingNgTomlinsonMartin} a comparison is made between
swarm robotics and other distributed networks such as mobile wireless
sensor networks (WSN), mobile and vehicular ad hoc networks (MANETs
and VANETs), multi-robot systems and software agents.  We refer the
readers to those references for more details.  Here we concentrate on
physical robots (hence the use of ``robots'' rather than ``agents'')
and \textit{emergent behaviour}, a feature that is not emphasised in
most other distributed networks.  We will examine the security of the
communications between robots, with a focus on preventing the
disruption of emergent behaviour.

\subsection{Stigmery and local sensing} 
\label{sub:stig-local}

We will rely heavily on peer-to-peer communications to secure emergent
behaviour, but we will firstly discuss these two communications methods
peculiar to swarm robotics.

Stigmergy refers to the communication of robots via the environment.
Robots modify the environment which in turns affect the behaviour of
other robots.  These messages left in the environment may “fade” over
time and eventually disappears.  This is a feature seldom discussed in
other types of network.  However, there are parallels in some
applications in other networks.  For example, one could consider the
``environment'' as a shared memory space such as a bulletin board.
This is used commonly in voting schemes (for example, \cite{FOO}).
``Fading'' messages can be modelled by using time-stamps or a
time-discount function.  For instance, in \cite{LiMalip}, a
time-discount function is used to ``fade'' a reputation.  All these
have to be achieved while guaranteeing the authenticity or integrity
or the messages.  Much of this have been discussed in
\cite{LaingNgTomlinsonMartin}.  

In some cases robot behaviour is determined not by direct
communication with other robots or central control, but by what the
robot can observe in its immediate surrounding, such as the
configuration or behaviour (such as proximity or velocity of movement)
of its neighbours.  Some networks have this feature.  For example, in
VANET, a vehicle may use the speed or location information of other
vehicles to determine its own behaviour.  Conventionally, when we
discuss security, we consider the confidentiality, integrity and
authenticity of the information that is exchanged.  Here the question
of confidentiality does not arise, if the robots are observable by
all. However, it is clear at least that the integrity and authenticity
of the information obtained by observation is tied in with the
legitimacy of the observed robots.  It was argued in
\cite{LaingNgTomlinsonMartin} that the authenticity of a robot cannot
be guaranteed unless there is some shared secret within the swarm.  We
will discuss later how we can obtain some assurance of whether a robot
is still legitimate or whether it is malfunctioning or has been
subverted.

\subsection{Emergent behaviour}
\label{sub:emergent}

The ``emergent behaviour'' aspect of swarm robotics presents one of
the greatest challenge to security.  Emergent behaviour arises from
local interactions amongst robots and between robots and the
environment.  (We do not concern ourselves here in how to design
individual behaviour in order to achieve the task at hand.  We are
concerned about the implication on communications security.) While it
seems possible to guard against external adversaries using
conventional peer-to-peer protocols of key distribution and
authentication, guarding against internal adversaries would appear to
be a lot trickier.  Robots are mobile and compromised robots may
spread their influence throughout the swarm, affecting local behaviours
which in turn may disrupt emergent behaviour.  How would one
distinguish between emergent and malicious or malfunctioning erratic
behaviour?  This is discussed in some depth in
\cite{LaingNgTomlinsonMartin}, arguing that traditional anomaly-based
and misuse-based intrusion detection methods do not work.  We will
consider how secure communications may be used to ameliorate the
situation.  It would appear that the policing of malicious/erratic
behaviour should not be too tight, so that some erratic behaviour is
still allowed in case it is in fact emergent.  This means that any
solution would have to be flexible so that it can be tuned by the
designer of the swarm.

\section{Types of swarm and threats} \label{sec:types}
 
\subsection{Classifying the swarm} \label{sub:class}

Before we describe our approach to securing swarm communications to
preserve emergent behaviour, we give a rough classification of robotic
swarms in terms of the homogeneity of its robots, the interaction
between robots, and the interaction with a central control.  We will
discuss the adversaries in the next section.

\begin{enumerate}
\item {\bf Homogeneity of robots.}  \label{robots}
  \begin{enumerate}
    \item \label{robots:hom} 
    Homogeneous robots.  

It is most generally accepted that the swarm is composed of a large group
or a few large groups of homogeneous robots.

    \item \label{robots:hie} Hieracrchical structure. 

It is possible that some
swarms may have ``special'' robots which are given a bigger role.
This role may be that of control (the ``master'' robot may direct
other robots to certain location or it may trigger an update of some
kind in the neighbouring robots), or it may be that of greater
capability (the ``master'' robot may have more keys so that it could
communicate with more robots), or it may be that of a sink - it
gathers all the data from its neighbours and reports back to a central
control.  (This is not an uncommon scenario in wireless sensor networks 
\cite{WSNsurvey}.)
  \end{enumerate}
\item {\bf Interaction between robots.} \label{rob-rob}

We assume that
there is no secret or hidden channels of communications, and that an
adversary may eavesdrop on or interfere with all communications.

  \begin{enumerate}
    \item \label{rob-rob:direct} Direct communication. 

This may be a broadcast, or 
it may be  a peer-to-peer communication.  The channel may be
a WLAN channel, a Bluetooth channel, an RFID channel, or it may use 
infrared or audio.  

    \item \label{rob-rob:env} Stigmergy. 

Robots communicate using stigmergy, that is, via the environment.
They leave messages for other robots by modifying the environment.   
This message will ``fade'' over time and eventually disappears.  

    \item \label{rob-rob:sensing} Local sensing. 

A robot may make decisions on how to behave by observing the physical
traits of its neighbours, such as their proximity, velocity, or
configuration.

  \end{enumerate}

\item {\bf Interaction with a central control.} \label{cc}

Again we make the same assumption as above that
there is no secret or hidden channels of communications, and that an
adversary may eavesdrop on or interfere with all communications.

  \begin{enumerate}
    \item \label{cc:none} No interaction between robots and central
      control after deployment.

This is commonly assumed, though it is not clear that it is entirely necessary,
desirable, or practical.

    \item  \label{cc:cc-rob} Central control to robots. 

It is sometimes accepted that there may be a central control
entity which can broadcast messages to the swarm.  These messages may
include key management messages or software updates.

This is generally a broadcast channel, though it is possible that in 
a more hierarchical swarm, the messages are relayed via the ``higher''
robots.

    \item \label{cc:rob-cc} Robots to central control. 

It is also possible that robots of the swarm send messages back to
central control.  This is required in some applications, such as that
of search and rescue.  This can be done either by beaconing, or by
relaying messages along to the sink (some designated reporting robot)
which is responsible for reporting back.

    \item \label{cc:twoway} Two-way communications between robots and
      central control.

Certain applications may require that robots and central control remain
in contact.  

  \end{enumerate}
\end{enumerate}

\subsection{Communications and computation capabilities}
\label{sub:ccc}

Computational abilities of robots vary.  We will assume that the
robots in this paper have enough capability to perform basic
cryptographic computations .  This is not unrealistic: advances in
hardware design and manufacturing has resulted in small embedded
devices that are capable of executing public key cryptography and
other complex computational tasks \cite{Ferrer}.  For example, even
the first generation Raspberry
Pi\footnote{https://www.raspberrypi.org/} provided a real-world
performance roughly equivalent to a 300 MHz Pentium II of
1997--99\footnote{https://en.wikipedia.org/wiki/Raspberry\_Pi}. The
robots used in \cite{Chamanbaz2017,Kit2019} are able to communicate at
a range of up to 300m.  Some swarm robots are large machines used for
precision agriculture and for these onboard computation is not a great
restriction.  At the other end of the size spectrum, we have
nanorobots.  For example, the design of an artificial mechanical red
blood cell or ``respirocyte'' envisions that a 104 bit/sec
nanocomputer meeting all its computational requirements, which is
roughly about 1/50th the capacity of a 1976-vintage Apple II
microprocessor-based PC \cite{nanorobots}.

\subsection{Threats to the swarm} \label{sub:threats}

We assume that the aim of an adversary is to disrupt the swarm.  It 
may do so by 
\begin{itemize}
\item discovering secrets and confidential information, or
\item impersonating or corrupting or introducing robots to masquerade
  as robots of the swarm to gather information, plant false
  information or to change the swarm's behaviour by its robots' own
  behaviour, or
\item removing robots or information from the system.
\end{itemize}

We will not discuss the first threat in detail, since it is discussed
in many other work (such as \cite{HTM,LaingNgTomlinsonMartin}), and
also we are more interested in what security can be vouchsafed in
publicly observable behaviours.  Instead we will focus on the second
and third threats, specifically to emergent behaviour.  We will
consider different classes of adversaries as follows.

\begin{enumerate}
  \item \label{adv:insider} Insider/outsider. (What they know.)

An adversary who is an outsider has no access to the cryptographic
keys and credentials of the robots.  This type of adversary and the
threats it poses and possible mitigation are discussed in depth in
\cite{LaingNgTomlinsonMartin}.  Most threats can be dealt with by
having some sort of secrets known only to the swarm, and both public
key cryptography and key predistribution for symmetric key
cryptography can be used to prevent any attacks.  It is not clear what
an outsider in the context of local sensing is - we will assume either that
it would simply look different and will be disregarded by robots of
the swarm, or it would fail some sort of authentication process.

Insiders, in contrast, have access to keys and credentials.  They may
be corrupted robots, or they may have keys and credentials
manufactured by an adversary. The use of threshold schemes and
intrusion detection systems is discussed in
\cite{LaingNgTomlinsonMartin}, and it is found that neither of these
solutions are ideal in the context of local sensing and emergent behaviour:  
apart from the issue of removing such an adversary, it is not easy to 
distinguish bad behaviour from emergent behaviour.

  \item \label{adv:active} Active/passive. (What they can do.)

A passive adversary eavesdrops and tries to deduce secrets and
information.  Using encryption would prevent an outsider from 
doing this but this is ineffective against an insider.  In the context of
local sensing, we assume that an insider passive adversary would simply
observe and possibly record the behaviour of the swarm robots.

An active adversary, in addition to eavesdropping, may modify or inject
messages, or participate in the swarm while behaving incorrectly.
Again, having some form of public key or symmetric key cryptography 
and using this to authenticate messages and robots would thwart an
active outsider.  What is more difficult to
address is an active insider behaving in a way to subvert the purpose of
the swarm.  This is the issue we will deal with in this paper.

  \item \label{adv:local} Local/global. (How many robots do they affect.) 

A local adversary can only affect the robots local to it.  It has no
veiw of robots not local to it.  
It does not know whether they are
communicating or what they are communicating, nor does it know their
behaviour or action.  A global adversary, on the other hand, sees the
behaviour of the entire swarm: the behaviour of the robots, and the
presence of communication between robots.

This class of adversaries appear to be different from the usual kinds
of adversaries that are discussed in the literature (\ref{adv:insider}
and \ref{adv:active}) and seems quite pertinent to swarm security.
Given that emergent behaviour arises out of local interactions, it may
be that a global adversary could coordinate disruptions at different
localities to effect a disruption of emergent behaviour more
efficiently.  Hence, as a defence, we would like to ensure that local
behaviours are not disrupted.

  \end{enumerate}

Since outsiders can mostly be dealt with using appropriate cryptographic
mechanisms we will focus here on insider adversaries.  Passive insiders
would appear to be hard to identify, and we will consider what can be done
when such adversarise are active.  In dealing with local disruptions we hope
to prevent them propagating to disrupt global emergent behaviour.

\subsection{Some examples}
\label{sub:examples}

We will consider some examples to illustrate the usefulness of this
classification of types of swarms and threats to the swarm.

Suppose we only expect adversaries who are passive outsiders, and suppose our
swarm consists of homogeneous robots (1(a) of Section \ref{sub:class})
with only direct communication between the robots
(2(a)).  Suppose there is no interaction
with central control (3(a)).  One simple solution
could be to equip each robot with a single encryption key, to prevent
eavesdropping.

Suppose however that the adversaries are active outsiders who might
attempt to manipulate messages, we could give each robot, in addition,
a single signing key and its certificate, issued by central control
prior to deployment. These keys may not expire during the event or the
action time.

Another possibility is to
deploy our robots with random key predistribution \cite{EG}.  There is
a fixed probability, decided upon before deployment, that two robots
can authenticate and communicate with each other. Such a solution is effective
again active outsider adversaries.  However, this does not allow
adaptation if the environment should change.  One may ask whether
changing one of the conditions might allow more flexibility.  Indeed,
if a broadcast channel from central control to the robots should be
made available (3(b)) then one could deploy the
broadcast enhanced key predistribution scheme proposed in
\cite{BEKPS}:
  \begin{enumerate}
  \item Key pre-distribution: each robot is given a set of underlying
    keys prior to deployment.  These keys are used only for the
    encrytion and decryption of temporal keys.
  \item Periodic broadcast from control to robots after deployment:
    send temporal keys for use in communication.  Temporal keys are
    encrypted using underlying keys so that a robot learns a temporal
    key only if the temporal key is encrypted by an underlying key
    known to the robot.  The distribution of temporal keys can be
    adjusted according to the desired connectivity and resilience at
    particular times.
  \item Robots discover common keys by broadcasting identifiers of 
  temporal keys.
  \end{enumerate}
In addition, if we choose the underlying key pre-distribution scheme
carefully, we would be able to revoke a robot if it is known to be
malfunctioning or captured.  This gives the swarm some resilience
against an active insider adversary. This example was discussed in the
``Further Research'' section of \cite{LaingNgTomlinsonMartin}.

It would be interesting to study solutions to specific types of swarm
and what might be adapted if certain conditions are tightened or
relaxed. 

\section{A scheme to protect emergent behaviour}
\label{sec:scheme}

\subsection{Overview}
\label{sub:overview}

Suppose that our robots are capable of public key cryptography, and
are given individual public and private keys.  If the public keys are
signed by the central control, this guards against external
adversaries even though it is not secure against internal adversaries.
Given that we can now authenticate a robot by its public key, how do
we know if a robot is not malfunctioning or corrupt?  If we know a
robot is malfunctioning or corrupt it can be revoked.  There are many
solutions to the revocation problem, including revocation lists.
However, how to decide whether a signer or a signing key should be put
into the revocation list is often the trickier problem . In this work,
we try to give one solution for how to identify and revoke a bad robot.

However, this on its own may not be sufficient.  If a malicious or
malfunctioning robot is not filtered out by the step above it could
still behave badly to affect its neighbours' behaviour.  What we can
do about this is to take some sort of consensus from neighbours.  For
example, if a robot observes a few neighbours, and one of them behaves
in a different way from the others, the robot could make a decision to
follow the majority with some probability and follow the minority with
some other probability. What these probabilities are would
be up to the engineer of the swarm. Alternatively, a robot could
consider another robot more trustworthy if it has encountered that
robot regularly in the past, and consider a robot less trustworthy if
it has not encountered it before, or only has a report of it from some
other robot.  Our goal is how to ensure that the information gathered
by a robot is trustworthy.

By identifying and revoking bad robots with high probability,
and preventing bad robots from having too much influence on local
behaviour, we can maximise the chances of emergent behaviour.

We assume that robots are capable of public key cryptography and that
they are equipped with a hardware clock synchronised before
deployment.  Clocks may drift and we allow some margin of error.  We
assume there is a central control which installs all the necessarily
credentials and algorithms before deployment and has no more contact
with the robots after deployement.  We also assume the presence of an
adversary who is an active insider with a global view, but we assume
that the fraction of corrupt and malfunctioning robots is small.

To start with we assume that robots are deployed in one single area
such that within some time interval $\Delta$ all robots would have
exchanged information with some other robot.  We consider a robot
suspicious - they might be malfunctioning or they might have been
tampered with - if they make false reports or if they are taken out of
the systems for a certain number of time intervals.  We aim to
identify this behaviour: when two robots meet they make a record of
the encounter, and exchange their history - here this means a record
of what robots they have met in a specified number of past time
intervals.  We will see that this simple mechanism allows us to
achieve our goal with high probability.

\subsection{The scheme}
\label{sub:thescheme}

We assume that central control has a signature scheme
$\scheme(\sign_{CC}, \ver_{CC})$.  Central control is assumed to be
trusted.  To set up a task for the swarm, it installs all the
necessarily credentials and algorithms in the robots before deployment
and has no more contact with them afterwards until the task is
completed.

There are $N$ robots $R_1$, $\ldots$, $R_N$.  Each robot $R_i$ has:
  \begin{itemize}
  \item a signature scheme $\scheme(\sign_{i}, \ver_{i})$, and a
    certificate $\cert_i$ from central control (so each robot has a
    unique verifiable identity associated with ($\ver_{i}, \cert_{i}$));
  \item the signature verification algorithm of the central control
    $\ver_{CC}$;
  \item a clock that is capable of measuring time intervals;
  \item a hash function $h$.  
  \end{itemize}

We assume that time is divided into intervals, and the first time
interval is $t=1$.  Each robot $R_i$ maintains a signed list for time
interval $t$, $\history^t_i$, and we set $\history^0_i=\emptyset$.
Within a time interval $t$ a robot $R_i$ may meet other robots.  When
it does it records the encounter in $\history^{t}_i$: if $R_i$ meets
$R_j$ they exchange their history from the previous time interval, so
$R_i$ gives $R_{j}$ the signed list $\history^{t-1}_i$, and $R_{j}$
gives $R_i$ the signed list $\history^{t-1}_{j}$.  They also exchange
their authenticated verification algorithms $(\ver_i, \cert_{i})$ and
$(\ver_j, \cert_{j})$.  Robot $R_{i}$ then checks the validity of
the signed list $\history^{t-1}_{j}$ it receives by verifying the
signature on it. Similarly $R_{j}$ checks $\history^{t-1}_{i}$.
We assume that a time interval is short enough that a robot
can only exchange history with its immediate neighbours once. 

At the end of
time interval $t$, $R_i$ constructs an event list $\event^t_i$:

\[ \event^t_i = \left\{ (R_{i_1},\history^{t-1}_{i_1}),  (R_{i_2},\history^{t-1}_{i_2}),  \ldots, (R_{i_k},\history^{t-1}_{i_k}) \right\}, \]

if $R_i$ encountered robots $R_{i_1}, R_{i_2}, \ldots, R_{i_k}$ in
time interval $t$, or $\event^t_i= \emptyset$ if it did not encounter
any other robots.  If any $\history^{t-1}_{i_j}$ is missing or
does not verify, then the
entry $(R_{i_j},\history^{t-1}_{i_j})$ is omitted.
A new history list is also constructed:

$$\history^{t}_i = \left\{ \event^t_i, t, \history^{t-1}_i,
\sign_i(h(\event^t_i, t, \history^{t-1}_i)) \right\}. $$ 

So each robot constructs a chain of events.  At the end of each time
interval $t\ge 1$ a robot constructs a link of the chain $(\event_i^t,
\history_i^t)$, where $\event_i^t$ contains information about events
that happened in the time interval $t$, and $\history_i^t$ links
$\event_i^t$ to events that happened in previous time intervals
described in $\history_i^{t-1}$.  In this way $R_i$ has a record of
all the robots it has met as well as the robots that these robots
claimed to have met.  An encounter between two robots $R_i$, $R_j$
in time interval $t$ is accepted if $R_i$ has $\history^{t-1}_j$
and $R_j$ has $\history^{t-1}_i$. 

Based on this record, $R_i$ can analyses the behaviour of each robot
to discover a ``bad'' robot, such as a robot that makes false reports
or that disappears for too long. The details of the analysis 
are given in the next subsection. After analysis, $R_i$ can put bad
robots into its local revocation list. The policy on what kind of
behaviour a robot has that leads to revocation would be decided by the
engineer of the swarm and this is out scope of this paper.

After that task of the swarm finishes, the history lists of each robot will
be collected by the central control and further analysis will be
made. The whole swarm system can benefit from the above information
collection during the job.  As this procedure is straightforward, 
we do not discuss it further in the paper.

(We note that robots in collusion may either simply share each other's
private keys, or create lists in advance and distribute them on each
other's behalf.  Hence a challenge/response protocol may not add more
security.  However, while they can always vouch for each other, they
can't pretend to have met honest robots, because these lists have to
be signed too.)

\section{Analysis of the above scheme}
\label{sec:analysis} 

We model the activities of the $N$ robots in any time interval $t$ as
vertices in a binomial random graph $G_t=G_t(N,p)$, with an edge
between vertices if the corresponding robots have exchanged
information within that interval, and this happens with probability
$p$.  We write $E(G_t)$ for the set of edges of $G_t$, and we write
$N_{G_t}(R)$ for the set of vertices that are neighbours of $R$ in
$G_t$, that is, the set of vertices that share an edge with $R$ in
$G_t$.  This gives all the robots that $R$ meets in time interval $t$.
The expected mean degree of a vertex is $Np$, and this gives the
expected number of robots a robot will meet in one interval (see
\cite{RandomGraphs}).

\begin{example}[Numerical example]
  \label{eg:numerical}
In \cite{Chamanbaz2017} an experiment
involving exploration using robots used $N=25$ robots in a $1km^2$
area, which gives robot density of about 1 in 40 000 $m^2$.  If we
consider the area as a $1km \times 1km$ square, subdivided into $200m
\times 200m$ grids, with a robot in the centre of each grid, then
since the robots were able to communicate at over $300m$, this allows a
robot to see the 8 robots in the neighbouring grids, out of 24 other
robots, giving $p=0.33$.  At the upper end of the experiment, $N=48$
robots were deployed in a larger area, giving $p=0.17$ using the same
assumptions.  We will run a rudimentary \textsc{Mathematica} program
with random graphs of these parameters to confirm our theoretical
analysis.
\end{example}

\subsection{Correctness}
We first discuss correctness of a swarm system and claim that a swarm
system described in Subsection~\ref{sub:thescheme} holds two
correctness properties, namely {\it System Correctness} and {\it Local
  Correctness}.

\subsubsection{System Correctness}
\begin{theorem}[System Correctness] Under the assumption that every robot of the swarm is honest, a full collection of all robots' individual history lists will be a true record of the communications among them.
\end{theorem}

\begin{proof} This property can be argued straightforwardly. Let $N$ be the total number of robots in the swarm and $T$ be the total number of intervals during a task. Following the scheme description of Subsection~\ref{sub:thescheme}, a history list made by robot $R_i$, $i \in \{1, 2, ..., N\} $ is:
$$\history^{T}_i = \left\{ \event^T_i, T, \history^{T-1}_i,
  \sign_i(h(\event^T_i, T, \history^{T-1}_i)) \right\}, $$ and a full
  collection of all robots' individual history lists are
$$\history^{T} = \history^{T}_1 \wedge \history^{T}_2 \wedge ... \wedge \history^{T}_N. $$  

  Because all robots are honest, every event where two robots, say
  $R_i$ and $R_j$, met each other must be recorded in
  $\history^{T}_i$ and $\history^{T}_j$, and each robot will correctly
  report every meeting that it was involved in.  More specifically, if there is a record of $R_i$ meeting $R_j$ in interval $t$ then $R_i$ did indeed meet $R_j$ in interval $t$ and there is a record of $R_j$ meeting $R_i$, and vice versa: if $R_i$ met $R_j$ in interval $t$ then there is a record in $R_i$ (and $R_j$)'s history. Records must be paired. There is neither any forged reports nor any missing reports.  Therefore, $\history^{T}$
  must be a true record of the communications among all the robots of
  the swarm. So the theorem follows.
\end{proof}

\subsubsection{Local Correctness}
\begin{theorem}[Local Correctness]
  Under the assumption that every robot of the swarm is honest and
  that the swarm communication patterns follow a binomial random graph as
  described at the start of Section \ref{sec:analysis}, after a
  certain number of time intervals, the probability that each robot's
  local history list will cover more or less the same information of
  any other individual robot's record, which is a true record of the
  communications of the whole swarm during these intervals, is significantly high.
\end{theorem}

\begin{proof}
  Because all the $N$ robots are honest and their communication
  patterns follow a random graph $G(N,p)$ in every interval, the
  probability that a robot does not meet another robot in $\Delta$
  intervals is $(1-p)^{(1+\frac{(\Delta-1)Np}{2})\Delta}$, as
  calculated in Section \ref{sub:disappearance}.  Since $1-p<1$, this
  probability tends to 0 quadratically with
  increasing $\Delta$.  Hence the probability that a robot has met all
  other robots after some $\Delta$ intervals is reasonably high.
  We can therefore assume that in every $\Delta$ continuous intervals,
  a robot will meet every other robot and exchanged its history list
  with them at least once. The robot then obtains a full collection of
  all robots' individual history lists up to the point of time when it
  is the beginning of these $\Delta$ intervals. Based on the
  discussion of system correction above, the robot should have a true
  record of the communications among all the robots of the swarm
  before that time. So the theorem follows.

  To illustrate this property, let us recall the numerical example in
  Example \ref{eg:numerical}, with $N=25$, $p=0.33$.  In every
  $\Delta=3$ continuous intervals, the probability that a robot does
  not meet another robot is $(2/3)^{28} \approx 1.49 \times
  10^{-5}$. Hence the probability that a robot has met all other
  robots in 3 intervals in either case is very high ($\approx
  0.99998$) and the probability of having a true record of the
  communications of the whole swarm during these 3 intervals is high.
\end{proof}

\subsection{Threat Model} 

A robot is bad if it makes false reports, or  if it disappears for too
long. We will analyse these two bad behaviours respectively in the
following two subsections.  We assume that a bad robot has the same
capability as a good robot.

Our goals are to prevent bad robots from having too much influence on local 
behaviour during the execution of the task, and to identify bad robots.  The first
goal, as discussed in Section \ref{sub:overview}, can be achieve by identifying as
many bad robots as we can during and after the execution of the task.  To do this,
we identify two suspicious behaviours:

\begin{enumerate}

\item If a robot disappears for too many intervals we suspect it of either malfunctioning or having been
captured and subverted.  We aim to identify such a robot.
\item A robot may make false reports in order to make a robot
distrust the good robots around it or to disguise its own or its fellow bad robots' status.  We aim to prevent framing of good robots and to detect collusion.
\end{enumerate}

We assume that a small fraction $\alpha$ of robots are corrupt or are
malfunctioning (``bad''), $0 < \alpha \ll 1$.

\subsection{Making false reports}
\label{sub:false} 

Suppose $R_i$ is a bad robot and wants to make a false report about an honest robot $R_j$.  
There are two types of false reports: 

\begin{enumerate}
\item $R_i$ claims to have met $R_j$ in time interval $t$ even though
  it has not.

    To claim this $R_i$ must prove that it has $\history^{t-1}_j$
    which contains $R_j$'s signature.  This cannot be done if the
    signature scheme is secure. 
    
    Also, if $R_j$ does not have $\history^{t-1}_i$ in its own $\history^{t}_j$, a record of the fake meeting between $R_i$ and $R_j$ at time interval $t$ will not be accepted by other honest robots later. To make the exchange record be paired and be accepted by others, $R_i$ also needs to let $R_j$ record $\history^{t-1}_i$. This cannot be done by $R_i$ itself.
    
    Another possibility is that at time interval $t' > t$, some
    other, possibly corrupt, robot $R_k$ passes a legitimate $\history^{t-1}_j$
    to $R_i$ and a legitimate $\history^{t-1}_i$ to $R_j$, but $R_i$ cannot incorporate $\history^{t-1}_j$ into its history without modifying the hash chains, and this can be detected by other robots.  Any attempt of $R_k$ to pass  $\history^{t-1}_i$ to $R_j$ at time $t$ will also be detected by $R_j$ when the signature is verified.

\item $R_i$ claims not to have met $R_j$ in time interval $t$ even though it
  has, in order to give the impression that $R_j$ is a
  suspicious robot and has disappeared for too long.

  In this case $R_i$ simply does not record $R_j$'s chain
  $\history^{t-1}_j$.  If $R_i$ is the only robot to meet $R_j$ then
  it can try to convince other robots that $R_j$ has disappeared.
  However, if the expected mean degree $Np$ of the random graph
  $G(N,p)$ is greater than 1 then it is likely that $R_j$ would have
  met $Np-1$ other robots, and if the number of rogue robots is small
  then it is likely that this sort of attack would not have any
  significant effect on $R_j$. 
  
  It is possible also that $R_i$ refuses to give $\history_i^{t-1}$ to
  $R_j$, but this would make it seem more likely that $R_i$ is
  suspicious, and would be the correct outcome.
  
  Indeed, if we assume that the proportion of bad robots $\alpha$ is small,
  and we consider a record of an encounter
  trustworthy if at least $(1-\alpha)Np $ of them are paired, 
  then this sort of attack would not succeed.
  
  {\bf Numerical example:} We consider the case when a proportion
  $\alpha$ of the robots do not record encounters with other robots.
  Our \textsc{Mathematica} program showed that, for $N=25$, $p=0.33$,
  $\Delta=3$, even with up to $\alpha = 1/3$ of the robots corrupt,
  there is no effect on other robots being seen or reported seen.  For
  $N=48$, $p=0.17$, all robots were reported seen in $\Delta=4$
  intervals with again $\alpha = 1/3$ of the robots corrupt ($\Delta =
  3$ if there are no corrupt robots).  This confirms the theory that
  as long as the mean degree of the random graphs is high enough, bad
  robots are not able to ``frame'' other robots.
  
\end{enumerate}

In the situation where both $R_i$ and $R_j$ are corrupt, we 
may assume that they are in collusion, and share each other's
    private keys.  In this case $R_i$ may construct a $\history^{t-1}_j$
    and claim that $R_j$ has been seen.  Indeed, robots in collusion
    may vouch for each other, even if they cannot fake uncorrupted
    robots' lists.

    However, we can then calculate the
    probability of a robot meeting another robot - $R_i$
    meets $R_j$ with probability $p$, so $R_i$ meets $R_j$ in all
    $\Delta$ time intervals with probability $p^{\Delta}$, and if we
    see that $R_i$ meets $R_j$ too often then we may assume they are in
    collusion.  

    {\bf Numerical example:} With $N=25$, $p=0.33$, and $\Delta = 3$
    intervals, this gives the probability of $p^{\Delta} \approx
    0.036$ for a pair of robots to meet in 3 time intervals.  With
    $N=48$, $p=0.17$, the probability is approximately $0.005$.  Hence
    the probability of a pair of robots meeting in many intervals is
    low, and if a pair of robots meet too often we may treat them as
    suspicious.

\subsection{Disappearance}
\label{sub:disappearance}

A robot also becomes suspicious if it disappears for too many
intervals.  For example, let us consider the probability that a robot $R$
gets a report of another robot $R'$ within, say, $\Delta=3$ time
intervals.  This \textit{does not} happen if
\begin{itemize}
\item $R$ does not meet $R'$ at $t=3, 2$ and 1, that is, $(R, R') \not\in
  E(G_3)$ and $(R, R') \not\in E(G_2)$ and $(R, R') \not\in E(G_1)$.  This has
  probability $(1-p)^3$.
  
\item None of the robots $R$ meets at $t=3$ has met $R'$ at $t=2$ or
  $t=1$, that is, $(R'', R') \not\in E(G_2) \cup E(G_1)$ for all $R''
  \in N_{G_3}(R)$.  This has probability $(1-p)^{2 |N_{G_3}(R)|}$.
\item None of the robots $R$ meets at $t=2$ has met $R'$ at $t=1$,
  that is, $(R'', R') \not\in E(G_1)$ for all $R''
  \in N_{G_2}(R)$.  This has probability $(1-p)^{|N_{G_2}(R)|}$.
\end{itemize}

Hence the probability that $R$ has a report of $R'$ withtin $\Delta=3$
intervals is $1-(1-p)^{3+2 |N_{G_2}(R)| + |N_{G_1}(R)|} = 1-(1-p)^{3+ 3Np}$
since $R$ is expected to have degree $Np$.
If this is
sufficiently high that means that a legitimate robot would have been
seen by some other robot in two time intervals with high
probability, and therefore a robot not seen in three time intervals
may be regarded as suspicious and blacklisted.  

In general the probability that $R$ has a report of $R'$ withint $\Delta$
intervals is $1- (1-p)^{\Delta + ((\Delta-1) + (\Delta-2) + \cdots + 1)Np}
= 1 - (1-p)^{(1+\frac{(\Delta-1)Np}{2})\Delta}$, which is increasing
with $\Delta$.

{\bf Numerical example:} Using the same experiment as described above
in Section \ref{sub:false}, with $N=25$ and $p=0.33$ this gives the
probability of 0.99998 for a robot having a report of another robot
within 3 time intervals.  With $N=48$ and $p=0.17$ the probability is
0.99403 within 3 intervals.  Our \textsc{Mathematica} program confirms
this.

\section{Conclusion and further work}
\label{sub:conclusion}

By modelling a swarm using random graphs, and ensuring that events are
recorded securely in a hash chain, we can allow robots to identify
``bad'' robots with a high probability, while ensuring that ``good''
robots were not adversely affected.  This goes some way towards
protecting emergent behaviour, by limiting the influence of bad robots
locally.

Note that here we have restricted ourselves to modelling the swarm as
a binomial random graphs.  In the situation where we desire the swarm
to aggregate or disperse we may consider sequences of graphs where the
number of edges increases or decreases through time, or where the
likelihood of having an edge between two vertices increases or
decreases depending on whether there is an edge previously.  It will
also be interesting to examine the properties of the swarm where bad
robots are modified to have additional capability, such as
broadcasting ability, or a higher communication range which may be
modelled as a subset of vertices with higher connectivity.  

\textbf{Acknowledgement}
The authors would like to thank Professor Stefanie Gerke, Mathematics Department, Royal Holloway, University of London, for her assistance in random graphs.

\bibliographystyle{plain}
\bibliography{swarmcomms.bib}

\end{document}